\documentclass[acmlarge,screen,nonacm,review=false,timestamp=false]{acmart}
\makeatletter
\newcommand{\confshort}{\acmConference@shortname}
\newcommand{\conffull}{\acmConference@name}
\newcommand{\confdate}{\acmConference@date}
\newcommand{\confloc}{\acmConference@venue}
\AtBeginDocument{
  \fancypagestyle{firstpagestyle}{
    \fancyhead{}%
    \fancyfoot[C]{}%
  }
  \fancyhf{}
  \fancyhead[LO]{\@headfootfont\shorttitle}%
  \fancyhead[RE]{\@headfootfont\@shortauthors}%
  \fancyhead[LE]{\@headfootfont\footnotesize \confshort, \confdate, \confloc}%
  \fancyhead[RO]{\@headfootfont\footnotesize \confshort, \confdate, \confloc}%
  \fancyfoot[C]{}%
}
\makeatother
\acmBooktitle{\conffull\@ (\confshort), \confdate, \confloc}

\AtBeginDocument{%
  }

\copyrightyear{2026}
\acmYear{2026}
\setcopyright{cc}
\setcctype{by}
\acmConference[FAccT '26]{The 2026 ACM Conference on Fairness, Accountability, and Transparency}{June 25--28, 2026}{Montreal, QC, Canada}
\acmBooktitle{The 2026 ACM Conference on Fairness, Accountability, and Transparency (FAccT '26), June 25--28, 2026, Montreal, QC, Canada}
\acmDOI{10.1145/3805689.3812346}
\acmISBN{979-8-4007-2596-8/2026/06}

\usepackage{cleveref}
\usepackage{bm}
\usepackage{delarray}
\usepackage{microtype}

\usepackage{enumitem}

\usepackage{graphicx}
\usepackage{tcolorbox} 
\usepackage{dirtytalk}
\usepackage{xspace}

\newcolumntype{H}{>{\setbox0=\hbox\bgroup}c<{\egroup}@{}} 

\newcommand{\negjudg}[1]{\textsc{\textcolor{magenta}{#1}}}
\newcommand{\posjudg}[1]{\textsc{\textcolor{cyan}{#1}}}
\newcommand{\negtok}[1]{\textsf{\textcolor{magenta}{#1}}}
\newcommand{\postok}[1]{\textsf{\textcolor{cyan}{#1}}}
\newcommand{\greendelta}{\textcolor{brown}{\Delta}}

\newcommand{\YesNo}{\texttt{Yes/No}\xspace}

\begin{document}

\title[Prompting from the bench: LLMs and ordinary meaning analysis]{Prompting from the bench: Large-scale pretraining is not sufficient to prepare LLMs for ordinary meaning analysis}

\author{Abhishek Purushothama}
\authornotemark[1]
\email{ap2089@georgetown.edu}
\affiliation{%
  \institution{Georgetown University}
  \city{Washington}
  \state{District of Columbia}
  \country{USA}
}

\author{Junghyun Min}
\authornote{Both authors contributed equally to this research.}
\email{jm3743@georgetown.edu}
\affiliation{
  \institution{Georgetown University}
  \city{Washington}
  \state{District of Columbia}
  \country{USA}
}

\author{Brandon Waldon}
\email{bwaldon@mailbox.sc.edu}
\affiliation{%
  \institution{University of South Carolina}
  \city{Columbia}
  \state{South Carolina}
  \country{USA}
}

\author{Nathan Schneider}
\email{nathan.schneider@georgetown.edu}
\affiliation{%
  \institution{Georgetown University}
  \city{Washington}
  \state{District of Columbia}
  \country{USA}
}

\renewcommand{\shortauthors}{Purushothama et al.}

\begin{abstract}
  In the U.S.\ judicial system, a widespread approach to legal interpretation entails \text{assessing} how a legal text would be understood by an \text{`ordinary' speaker} of the language. Recent scholarship has proposed that legal practitioners leverage large language models (LLMs) to ascertain a text's ordinary meaning. But are LLMs up to the task? As textual interpretation questions arise in spheres ranging from criminal law to civil rights, we argue it is crucial that models not be taken as authoritative without rigorous evaluation. This work offers an empirical \text{argument} against LLM-assisted interpretation as recently practiced by legal scholars and \text{federal judges}, who reasoned the large amount of data that models see in training would enable models to illuminate how people ordinarily use certain words or phrases. 
    In controlled experiments, we find failures in robustness which cast doubt on this assumption and raise serious questions about the utility of these models in practice. For the models in our evaluation, slight changes to the format of a question can lead to wildly different conclusions---a vulnerability that parties with an interest in the outcome could exploit. 
    Comparing with a dataset where people were asked similar legal interpretation questions, we see that these models are at best moderately correlated to human judgments---not strong enough given the stakes in this domain.
\end{abstract}


\maketitle

\section{Introduction}
\label{sec:intro}
Legal decisions often come down to the interpretation of written text (e.g., in a statute or contract). Usually such  interpretation is straightforward.
At times, however, the text is appreciably imprecise or ambiguous, leading to disputes about how to apply the text to a particular set of circumstances.
In the U.S.\ legal system, 
judges often place considerable weight on the `ordinary meaning’ (meaning as would be understood by ordinary speakers of American English) of a legal term \citep{book_ordinary_meaning_2016}.\footnote{In this paper, we are officially agnostic whether `ordinary meaning' is a useful legal analytical construct (or even a coherent linguistic concept). Our focus is on whether LLMs are useful for pursuing interpretative methodologies that are purportedly grounded in `ordinary meaning.'}

One recent case (\textit{Snell v.~United Specialty Insurance Co.}) 
involved interpretation of the term `landscaping' (see  \cref{fig:landspacing-example}).
A second case (\textit{U.S.~v.~Deleon}) turned on the interpretation of the phrase `physically restrained' and whether the phrase describes indirectly restricting movement by threatening someone with a gun\footnote{We provide some additional cases related to other areas of law in \cref{sec:appendix-other-cases}.}. 

How can a judge ascertain the `ordinary meaning' of a legal text? Often, the judge will deploy armchair intuition buttressed by hypotheticals and dictionaries \citep{krishnakumar2024textualism}. On occasion, judges reference corpora \cite{solan2017corpus, gries2018ordinary} and surveys \cite{tobia_new_methods_2024}. 

\begin{figure}[t!]
\centering\small
\begin{tcolorbox}\sf
John is a contractor with insurance that covers property loss, damage, or personal injury claims that arise due to his `landscaping' work.
\\
\hspace*{1.5em}John is employed by a family, the Smiths, to install an in-ground trampoline in the family’s backyard. A few years after John completes the project, the Smiths successfully sue John for injuries that their daughter sustained while playing on the trampoline. John files a claim with his insurance company to recover losses incurred from the lawsuit.
    \end{tcolorbox}
    Considering just how ``landscaping'' would be understood by ordinary speakers of English, is John covered by the insurance—yes or no?
    \caption{A legal interpretation scenario represented as a QA task with binary questions. The example is based on the case \href{https://media.ca11.uscourts.gov/opinions/pub/files/202212581.pdf}{\emph{Snell v.~United Specialty Insurance Co.}}\ and constructed in the style of our task.}
    \label{fig:landspacing-example}
    \Description{Figure 1. Fully described in the text.}
    \vspace{-10pt}    
\end{figure}

Enter LLMs. The advent of LLMs has produced considerable excitement in the legal field including for legal text interpretation.
\citet{hoffman_generative_interpretation_2024} opined that LLMs are valuable resources for ascertaining the ordinary meaning of legal text. 
Federal judicial opinions \citep{snellvsunited,unitedstatesvsdeleon} have included `direct queries' to LLMs on the ordinary meaning of `landscaping' and `physically restrained'\footnote{In \textit{Snell} and \textit{Deleon}, Judge Kevin Newsom posed the two following direct queries, respectively: ``\textit{What is the ordinary meaning of landscaping?}''\ and ``\textit{What is the ordinary meaning of physically restrained?}''. See \citet{waldon_llms} for a detailed discussion of the direct query method.} on the grounds that these models supposedly capture and can articulate patterns of ordinary language use.

These developments have been accompanied by a growing body of academic scholarship assessing the benefits, risks, and impact of these technologies from the standpoint of both legal theory and practice. The valence of this work ranges from critical to optimistic. The critical end of the spectrum includes recent evidence that LLMs’ interpretive judgments are highly sensitive to how queries about the meaning are phrased, suggesting that LLMs are not reliable tools for legal interpretation \cite{waldon_llms, choiOfftheShelfLargeLanguage2025}. Recent scholarship on the optimistic end of the spectrum suggests, by contrast, that LLMs provide a consistent and accurate window into `ordinary' speakers' linguistic intuitions \cite{hoffman_generative_interpretation_2024, martinezComputationalCanons2025}. 

The question of how to establish the `ordinary meaning' of an expression is a fraught one in legal theory~\cite{book_ordinary_meaning_2016,tobia2022ordinary, tobia2020testing}. However, the scale of data that LLMs are trained on, and the fluency of LLM-generated text, have sparked much interest among lawyers and judges~\cite{snellvsunited, hoffman_generative_interpretation_2024,engel2024asking}.~Optimists conjecture that LLMs may prove useful for empirical ordinary meaning analysis, providing more objectivity and comprehensiveness than any individual could. But first, LLMs must be  carefully evaluated to establish that they can deliver on this promise.

Hence, evaluations of LLM-based ordinary meaning analysis must consider:
\begin{enumerate}
    \item whether LLMs---which appear to be able to answer natural language questions but are subject to unexpected or unpredictable \textbf{sensitivity} to inputs\footnote{We discuss previous works demonstrating these effects in \Cref{sec:background-prompt-sensitivity}.}--- are also sensitive to the framing of the question for this task; and
    \item whether LLM judgments are \textbf{sound} insofar as they approximate the interpretive judgments of lay human readers~\cite{bystranowski2025statutory}\footnote{Consequently, using an LLM specialized to the legal domain or evaluating it with legal experts may be desirable in some circumstances, but would not be in line with `ordinary meaning' inquiry, which seeks to probe the understanding of the public writ large.}.
\end{enumerate} 

\noindent We offer these as \emph{necessary} conditions for real-world use, such that failure to meet either one should disqualify LLMs for ordinary meaning analysis. LLMs (or alternative systems) that meet these two criteria might still suffer from other critical drawbacks that render them unsuitable for the task. Both proponents and detractors of LLMs agree that empirical data---particularly surrounding the methodological robustness of LLM-assisted analysis and the correspondence of LLM outputs to ordinary human intuition---can help to advance this debate.

To that end, we conduct a large-scale\footnote{Our study significantly expands the body of evidence that informs the debate of LLM legal interpretation. Previous work has mainly provided qualitative arguments rather than full-fledged evaluation: \Citet{waldon_llms} test LLMs on just two interpretive scenarios, and \citet{choiOfftheShelfLargeLanguage2025} utilizes just five.} investigation into the stability of LLM-based legal interpretation methods, with a focus on LLMs' ability to produce ordinary meaning analyses consistent with human native speaker intuitions.
We consider whether LLMs' training data \emph{scale}, coupled with the ability to fluently navigate \emph{question--answer style interactions}, translates to the capacity for robust ordinary meaning analysis. Specifically, we investigate whether large-scale pretraining and instruction tuning are sufficient to yield systems that reliably provide sound answers to queries about ordinary meaning. Our study includes GPT-4, a platform system, as a point of reference.

A growing body of legal and NLP research on the linguistic capabilities of LLMs (reviewed in \cref{sec:background}) informs our two hypotheses:
(1)\label{ref:hypothesis-one}
LLM judgments are highly sensitive to subtle manipulations in how ordinary meaning queries are posed to the model.
(2)\label{ref:hypothesis-two}
LLM judgments are poorly correlated to human judgment.

To test our hypotheses, we create queries based on a previously developed set of 138 scenarios that assess linguistic interpretation in a variety of hypothetical insurance contract disputes \cite{waldon_vague_contracts_2023}, and query 14 open-weight models and one closed-weight platform model, across 9 systematic question variants~(\cref{sec:methods}).
We find in~\cref{sec:results-analysis} that LLM judgments are inconsistent across model family and size, as well as across choices in question phrasing. The outputs of some instruction-tuned models of specific sizes are correlated to human judgment only in some question variants; moreover, neither decoded tokens nor probability distributions offer a reliable source of human-like `ordinary meaning' judgments. Though LLMs possess undeniably fluent generation capabilities, our results add to a growing chorus of skepticism about LLMs as tools for legal interpretation~(\cref{sec:related-work}).

Our paper provides the most extensive experimental evaluation of LLMs for legal interpretation to date.\footnote{We have documented specific limitations of the work in \Cref{sec:appendix-limitations}.} We extend previous results on LLM prompt sensitivity to fresh data in legal interpretation. We additionally present an analysis of human correlation using existing human judgment data. Our rigorous experimental and analytical formulation lays the foundation for the evaluation work necessary to establish whether LLMs can be considered trustworthy tools for legal interpretation. We make the code, results, and analysis available publicly\footnote{Code is available in the repository:~\url{https://github.com/bwaldon/llms-legal-interp}}.

\section{Background}
\label{sec:background}

Legal interpretation is pervasive in the courts, reaching far beyond a single judge, court, or area of law.
Often, judges place considerable weight on \emph{ordinary meaning} in deciding the proper interpretation of the text.\footnote{That is, U.S. judges are often inclined to consider how the text would be understood by nonspecialists. As U.S Supreme Court Chief Justice John Roberts recently put it: \say{So the most probably useful way of settling all these questions [about ordinary meaning] would be to take a poll of 100 ordinary -- ordinary speakers of English and ask them what it means, right?} (Oral argument, \emph{Facebook Inc. v.~Duguid et al.}~\cite{scotusOralArguments2020FacebookvDuguid}.)}

While such analyses have previously relied on hypotheticals, dictionaries, corpora, and surveys \citep{krishnakumar2024textualism, waldon2024reading, gries2018ordinary, solan2017corpus, tobia_new_methods_2024}, in the current era of generative AI enthusiasm, some scholars have expressed optimism that LLMs will revolutionize legal interpretation.
Multiple variations of this idea have been put forward~\cite{hoffman_generative_interpretation_2024,engel2024asking,snellvsunited,unitedstatesvsdeleon,martinezComputationalCanons2025}.

\label{sec:background-prompt-sensitivity} One such claim is that LLM-based methods need not be perfectly accurate to be useful; they just have to be more robust to misuse than alternatives~\citep{hoffman_generative_interpretation_2024}. However, LLMs have been shown to be sensitive to prompts at multiple levels. \citet{sclar_quantifying_prompt_sensitivity_2024} showed how spurious character-level features can lead to large changes in model performance, while \citet{pezeshkpour-hruschka-2024-large} showed that the order of options can affect model performance in multiple choice question answering.~\citet{zhuo-etal-2024-prosa} have organized such prompt sensitivity into a dataset-independent metric, but use LLM probabilities as measures of confidence. Turning to legal interpretation, \citet{choiOfftheShelfLargeLanguage2025} tested the robustness of LLMs by generating 2,000 paraphrases for each of the 5 scenarios from \citet{hoffman_generative_interpretation_2024}, expanding from 20 machine generated paraphrases. \citet{choiOfftheShelfLargeLanguage2025} found evidence of considerable prompt sensitivity across various models, training methods, and output processing methods.~\citet{waldon_llms} similarly demonstrated the ease with which an LLM judgment could be manipulated through subtle prompt editing. Taken together, this prior work establishes a need for empirical analysis of how known biases and sensitivities affect LLM legal interpretation. Below, we rigorously test LLMs in legal interpretation in a large set of scenarios, using systematically controlled prompt variants.

\label{sec:background-omniscience}
A second line of reasoning holds that for hard cases that turn on ordinary textual meaning, LLMs are potentially \emph{better} than humans (and superior to existing tools) because they have been trained on so much ordinary English data (we call this the `omniscience' argument). Judge Kevin Newsom, in a concurring opinion in \emph{Snell} (the `landscaping' case), gives voice to this idea, citing the diversity of ordinary usage that these models might draw upon:
\say{models train on a mind-bogglingly enormous amount of raw data \dots as I understand LLM design, those data run the gamut from the highest-minded to  the lowest, from Hemingway [sic] novels and Ph.D.~dissertations to gossip rags and comment threads. Because they cast their nets so widely, LLMs can provide useful statistical predictions about how, in the main, ordinary people ordinarily use words and phrases in ordinary life} \cite{snellvsunited}.
He appears to tone down the omniscience argument in a subsequent case, saying LLMs \say{may well serve a valuable auxiliary role as we aim to triangulate ordinary meaning,} complementing \say{traditional interpretive tools [such as] dictionaries} \cite{unitedstatesvsdeleon}.

Is it plausible that a careful judge, armed with established tools of ordinary meaning analysis and a bevy of law clerks, would arrive at a better-informed result because an LLM was consulted?
Newsom's rationale is predicated on certain assumptions about LLMs that have been called into question \citep{waldon_llms,pruss-25}.
In particular, his comment about the vast training data behind LLM chatbots suggests he believes systems are capable of conducting  \emph{metalinguistic reasoning}\footnote{Metalinguistic reasoning is the ability to not only use language, but also explicitly reason about the use of language~\cite{begus25metalinguisticabilities}, distinct from conventional reasoning that describes `the general human capacity for truth-seeking and problem solving'~\cite{proudfoot2009routledge}.}~about the language in their training data in order to synthesize an interpretive conclusion. But recent scholarship casts doubt on the idea that LLMs are capable of deep metalinguistic reflection \cite{behzad-etal-2023-elqa, thrush-etal-2024-strange, cheng-amiri-2025-linguistic, begus25metalinguisticabilities}.
Instead, it seems likely that LLMs are good at imitating or summarizing metalinguistic text in the training data---be it dictionary definitions, textbooks, or online language forum discussions \cite{behzad-etal-2023-elqa, waldon_llms}. Thus, to the extent that an LLM produces plausible-sounding responses to an interpretive prompt, it likely draws on what humans \textit{say} about language \cite{almeman-etal-2024-wordnet}, rather than what they \textit{do} as language users.
Still, if LLMs could accurately synthesize ordinary people's opinions about meaning, then they might be a reliable and cost-effective tool for ordinary meaning analysis, as hypothesized by \citet{hoffman_generative_interpretation_2024}.

It is crucial, then, to test whether LLMs' interpretive judgments are reliably correlated to (ordinary) human interpretations. By some measures, LLM probability models exhibit human-like sensitivity to grammatical phenomena in English sentences: these include linguistic or syntactic acceptability~\citep{gauthier-etal-2020-syntaxgym,warstadt-etal-2019-neural,warstadt-etal-2020-blimp-benchmark}, and semantic plausibility \citep{kauf2023event, kauf-etal-2024-log}. However, there is increasing `behavioral' evidence of differences between LLMs and humans when it comes to learning and processing language \citep{oh-etal-2024-frequency, aoyama-wilcox-2025-language, mccoy2025modeling} and answering questions \cite{srikanth-etal-2025-questions}. To test whether LLMs and humans arrive at similar interpretive conclusions, we use a dataset of human judgments reported by \citet{waldon_vague_contracts_2023} to evaluate LLMs' interpretive judgments.

\section{Legal Interpretation with LLMs}
\label{sec:methods}
Consider the landscaping example in \cref{fig:landspacing-example}. A term within the contractor's insurance contract, `landscaping,' must be interpreted to determine whether the insurance covers the described scenario. The task explicitly demands a judgment as to how ordinary speakers would understand the contract language in context. 
This judgment may not cohere with one's beliefs regarding the `correct' interpretation of the provision, or regarding how a judge will actually resolve the legal dispute at hand.\footnote{Our investigation is perspectivist, as we are not assessing model behavior against a single, fixed `ground truth' linguistic interpretation or judicial outcome \citep{frenda2024perspectivist}.} Our assessment metrics assume that `ordinary interpretation' is both highly varied and to some extent subjective. Our first metric---robustness to variation---evaluates the stability of model judgments across multiple prompt formulations. Our second metric---human correlation (see \cref{sec:human-correlation})---evaluates the extent to which model judgments cohere with the intuitions of a relevant human population. In the remainder of this section, we describe a study designed to investigate LLMs' legal interpretation capabilities as assessed against these two metrics.

\paragraph{Materials}
\label{sec:method-vague-contracts}
Our study adapts materials originally developed by \citet{waldon_vague_contracts_2023} for a human study of legal interpretation and consists of 138 items based on real-world insurance contracts. \Cref{tab:prompt-structure} provides an example item. Each item names a category of insurance coverage (e.g., Vehicle Damage) and provides a definition of that category. The item then describes a policyholder's loss, which may or may not be covered by the named category.~\Citet{waldon_vague_contracts_2023} analyzed human responses from 1,338 U.S.-based native English speakers recruited via Prolific. The participants were shown a vignette composed of the scenario and insurance text. They were then asked to determine whether the vignette's protagonist (e.g., Ken) was covered by the insurance, with three response options: \posjudg{Yes}, \negjudg{No}, or \textsc{Can't Decide}. A sample vignette (\cref{fig:vague-contracts-vignette}) can be found in \Cref{sec:appendix-vague-contracts}.

\begin{table*}[ht!]
    \centering
    \small
    \begin{tabular}{ l p{0.7\linewidth}}
    \toprule
    \textbf{Query element} & \textbf{Example (`Vehicle Damage')} \\
    \midrule
Insurance text & Steve's car insurance policy includes coverage for ``Vehicle Damage,'' defined as ``loss or damage to the policy holder's 1) car; or 2) car accessories (while in or on the car)''  \\
         Scenario & One day, Steve is involved in a minor accident. His GPS navigation system, which was in the car at the time, was damaged. Steve files a claim with his insurance company for the damage. \\
         Framing for `ordinary meaning' & Considering just how ``accessory'' would be understood by ordinary speakers of English,  \\
    Question & 
      is Ken covered by the insurance—yes or no?\\
     Cue & Final answer is: \\
     \bottomrule
    \end{tabular}
\caption{Elements of the interpretive queries. `Insurance text' and `Scenario' are sourced from the study item, and the `Framing' and `Question' are added to solicit interpretive judgment. The `Cue` was found with exploratory tests to induce high rate of first token judgment.
}
\label{tab:prompt-structure}
\vspace{-10pt}
\end{table*}

\paragraph{Interpretation as binary QA}
\label{sec:binary-interp}
Following the design of the human experiments, we query LLM interpretation with binary questions.
As illustrated in \cref{tab:prompt-structure}, we frame polar judgment in legal interpretation as binary QA, where cues\footnote{Explicit cues (e.g. ``Is Ken covered by the insurance---yes or no? Final answer is:'') are likely different from how judges may use LLMs. But for empirical evaluation, we believe our \texttt{Yes/No} variant (\cref{tab:question-variations})~ is able to characterize a wide range of felicitous responses to how judges actually use LLMs---likely with simple questions (e.g. ``Is Ken covered by the insurance?'').} constrain the output space (e.g., that it should answer directly with ``yes'' or ``no''). The LLM judgment is operationalized as the first output token probability.
Our design allows us to obtain probabilities which can be used for more robust analysis, including correlation to human judgment.
It is also robust against incongruent text output, where the output string does not have a clear response.
Motivated by prior work suggesting a lack of alignment between token probability and intended model judgment \cite{wang2024look}, we manually compare our operationalization of LLM judgment as first token probability to manual extraction from decoded output and show that our operationalization is more robust (\Cref{sec:token-output-alignment}).

Additionally, we consider several types of \postok{yes} and \negtok{no} tokens across casing as indicators of coverage judgment, and operationalize their respective sums as judgment probabilities, which we further describe in \Cref{sec:judgment-as-sum}.
\label{sec:distribution-not-confidence}
However, we do not assume a direct association between token probability and judgment confidence.
This is true of our robustness analysis and human correlation analysis. For human correlation analysis, we specifically motivate a linking hypothesis detailed in~\cref{sec:human-correlation}.

\paragraph{Formulation of prompt variants}
To investigate whether LLMs' interpretive judgments are robust to minor changes in prompt design, we constructed a template of 9 question variants and applied that template to each of the 138 items. An example paradigm is presented in \cref{tab:question-variations}.

\begin{table*}[!ht]
    \small
    \centering
    \begin{tabular}{l p{0.8\linewidth}}
    \toprule
    \textbf{Variant} & \textbf{Question} \\
    \midrule
    & Considering just how the word `landscaping' would be understood by ordinary speakers of English,\\
    \\
    \texttt{Yes/No}  &is John covered by the insurance—yes or no?
    \\
   \texttt{No/Yes}
   & is John covered by the insurance—no or yes?
    \\
    \texttt{Negation} &  is John not covered by the insurance—yes or no? Final answer is:
    \\
    \texttt{Agreement} &  do you agree with the statement, “John is covered by the insurance”—yes or no?:\\
    \texttt{AgrWithNeg} &  do you agree with the following statement: “John is not covered by the insurance”— yes or no? \\
    \texttt{Disagreement} &  do you disagree with the following statement: John is covered by the insurance”— yes or no?\\
    \texttt{DisagrWithNeg} &  do you disagree with the following statement: John is not covered by the insurance”— yes or no?\\
    \texttt{Options} &  is John covered by the insurance?  Options: A. John is covered. B. John is not covered.\\
    \texttt{OptionsFlipped} & is John covered by the insurance?  Options: A. John is not covered. B. John is covered.\\
    \bottomrule
    \end{tabular}
    \caption{Systematic variation of the question (\cref{tab:prompt-structure}) in the interpretive queries. 
    See further discussion in \Cref{sec:question-variations-appendix}.
    }
    \label{tab:question-variations}
\end{table*}
\label{sec:methods-robustness}

Some variants reflect phenomena that are already attested to be challenging for LLMs. \citet{garcia-ferrero-etal-2023-dataset} and \citet{truong-etal-2023-language} show, for example, that LLMs find natural language negation words challenging and lack a deep understanding of the phenomenon. 
Moreover, \citet{sharma2024towards} and \citet{hong-etal-2025-measuring} demonstrate that in some contexts, LLM outputs are modulated by prompts that overtly solicit agreement (e.g., \textit{Do you agree that\dots ?}). Our study builds upon these previous findings in the domain of legal interpretation.

For most variants, an affirmative \postok{yes} response would correspond to the \posjudg{Covered} judgment, but in some  variants (e.g., the  \texttt{Negation} variant in \cref{tab:question-variations}), the \posjudg{Covered} judgment would be expressed with a \negtok{no} token. For clarity, we report and discuss probabilities corresponding to \posjudg{Covered} and \negjudg{NotCovered} judgments, rather than discussing \postok{yes} and \negtok{no} token probabilities. \label{sec:methods-double-negation}
We note that the \texttt{DisagrWithNeg} question variant is complex even for humans, as both negation words and negative prefixes add to the cognitive load \citep{SHERMAN1976143, Farshchi15012021
,Schiller07022017}. We still include the variant to evaluate whether language models are affected by such increased syntactic and semantic complexity and to assess their robustness in handling convoluted equivalence in prompt variation.

\paragraph{Models}
\label{sec:models}
Although our evaluation is focused on robustness and human correlation, our choice of models (i.e., exclusively pretrained or minimally post-trained models) isolates a theoretically relevant subclass of LLMs given the state of legal discourse around the use of LLMs for legal interpretation. In addition to pretraining and instruction tuning, various methods have been employed to improve LLMs in specific directions, including aligning LLMs to generate responses closer to \textit{human} preferences~\cite{ziegler2020finetuninglanguagemodelshuman, rlhf2026lambert_ch2}, training LLMs to follow procedural (e.g., chain-of-thought) steps~\cite{chung-jmlr-2024}, and applying inference meta-algorithms~\cite{welleck2024from}. However, pretraining is still the primary stage of LLM development, with instruction tuning providing the next distinct general purpose stage.

Hence, our model selection includes both `base' (pretraining only) and instruction-tuned models, of varying sizes up to 70B parameters, and GPT-4~\footnote{This is a closed weight model, whose judgment was obtained via API. See \Cref{sec:openai-appendix} for details.}. 
They include \texttt{Llama}~\citep{grattafiori2024llama3herdmodels}, \texttt{OLMo}~\citep{groeneveld-etal-2024-olmo}, \texttt{Mistral} \citep{karamcheti2021mistral}, and \texttt{Gemma} \citep{team2024gemma} and, we include \texttt{GPT-2} \citep{radford2019language} both as a model representing the smaller end and one that can support future careful investigation into how pretraining methodologies influence performance on this task.\footnote{In future work, we also plan to investigate the extent to which the ordinary meaning judgments of such models reflect generalizations of language use as observed in the models' pre-training data (or, e.g., reflect metalinguistic \textit{mentions} in that data).} The full list is in \cref{tab:models}, and implementation details in \Cref{sec:implementation-details}. Given our evaluation data's release in \citeyear{waldon_vague_contracts_2023}, data contamination is a possibility, which we discuss in \Cref{sec:appendix-contamination}.

As discussed in \cref{sec:background}, ordinary meaning analysis asks how language would be understood by a lay speaker of English---not a legal expert such as a lawyer or a judge.  We thus evaluate general-domain LLMs rather than models specialized for legal text and legal tasks (e.g., LegalBERT~\citep{chalkidis-etal-2020-legal}).

\begin{table}[tbh]
    \centering
    \small
    \begin{tabular}{l l}
        \toprule
        \textbf{Family} & \textbf{Models} \\
        \midrule
        Llama-3 & \texttt{1B}, \texttt{1B-Inst} (3.2), \texttt{3B}, \texttt{3B-Inst} (3.2), \texttt{8B}, \texttt{8B-Inst}   (3.1), \texttt{70B} (3.1), \texttt{70B-Inst} (3.3)\\
        GPT & \texttt{GPT-2-medium}, \texttt{GPT-4} \\
        OLMo-2 & \texttt{7B}, \texttt{7B-Inst}\\
       Ministral & \texttt{8B-Inst}\\
        Gemma & \texttt{7b}, \texttt{7b-it}\\
        \bottomrule
    \end{tabular}
    \caption{Our experiments include 15 models across 5 families: all are pretrained, some are instruction-tuned, and span range of sizes. GPT-2 is the smallest, a reference model on the low end (for considerations of modified pretraining investigations) and GPT-4 is a closed reference model on the other end (and has gone through additional stages such as alignment).}
    \label{tab:models}
\end{table}

\section{Results and Analysis}
\label{sec:results-analysis}
From the models, we collect both categorical (\posjudg{Covered} or \negjudg{NotCovered}) and distributional (probabilistic) judgments over tokens that represent the \posjudg{Covered} judgment (\postok{yes}, \postok{Yes}, \postok{YES} for \texttt{Yes/No} variant), tokens that represent the \negjudg{NotCovered} judgment (\negtok{no}, \negtok{No}, \negtok{NO} for \texttt{Yes/No} variant),
and the residual \textsf{\textcolor{gray}{other}} tokens. We use these judgments to analyze how robust the models are with respect to variation across family, size, and question. Additionally, we examine correlation (or lack thereof) between human and LLM judgment.

\subsection{Analysis of judgments for \YesNo prompts}

We begin by focusing on what is arguably the most basic question variant: \YesNo, illustrated in \cref{tab:prompt-structure}.
The categorical and distributional judgments for all models with the \YesNo question variant are reported in \cref{tab:judgment-distributions}.

\begin{table}[tbhp]
\centering
\small
\setlength{\tabcolsep}{3.5pt}
\begin{tabular}{@{}lcccc|lcccc@{}}
\toprule
& \multicolumn{2}{c}{\textbf{Categorical}} & \multicolumn{2}{c|}{\textbf{Distributional}} & & \multicolumn{2}{c}{\textbf{Categorical}} & \multicolumn{2}{c}{\textbf{Distributional}} \\
& \multicolumn{2}{c}{\textbf{Counts}} & \multicolumn{2}{c|}{\textbf{Spread}} & & \multicolumn{2}{c}{\textbf{Counts}} & \multicolumn{2}{c}{\textbf{Spread}} \\
\textbf{Model} & \posjudg{Covered} & \negjudg{NotCovered} & Min & Max & \textbf{Model} & \posjudg{Covered} & \negjudg{NotCovered} & Min & Max \\
\midrule
Llama-70B   & 28  & 110 & 0.21 & 0.48 & OLMo-2-7B         & 70  & 68  & 0.19 & 0.56 \\
\quad +Inst & 59  & 79  & 0.02 & 0.85 & \quad +Inst       & 53  & 85  & 0.00 & 0.99 \\
Llama-8B    & 80  & 58  & 0.14 & 0.37 & Ministral-8B-Inst & 75  & 63  & 0.21 & 0.59 \\
\quad +Inst & 0   & 138 & 0.11 & 0.74 & gemma-7b          & 39  & 99  & 0.19 & 0.49 \\
Llama-3B    & 127 & 11  & 0.09 & 0.52 & \quad +it         & 131 & 7   & 0.00 & 1.00 \\
\quad +Inst & 53  & 85  & 0.16 & 0.69 & GPT-4             & 77  & 61  & 0.00 & 1.00 \\
Llama-1B    & 138 & 0   & 0.06 & 0.29 & GPT-2-medium      & 5   & 133 & 0.13 & 0.31 \\
\cmidrule(lr){6-10}
\quad +Inst & 138 & 0   & 0.15 & 0.59 & Human Majority    & 84  & 54  & --   & --   \\
\bottomrule
\end{tabular}
\caption{Count and probability range for each model's \posjudg{Covered} and \negjudg{NotCovered} judgments in response to \YesNo questions. Both distribution and the effect of instruction tuning vary significantly across models. Human majority is also provided for reference.}
\label{tab:judgment-distributions}
\end{table}

\paragraph{Some models behave like stopped clocks.} We observe that many models repeat the same categorical judgment across all scenarios, or are highly biased towards one judgment: 6 of the 15 models tested
provide the same response to \YesNo questions representing more than 127 (92\%) of the 138 tested scenarios.
Three models provided the same judgment for all 138 scenarios.
In these cases, it is doubtful that the model judgment reflects substantive engagement with the provided scenario.
These models, rather, are only as useful as the proverbial stopped clock that correctly tells the time twice a day.


\paragraph{Different distributions for each model.} We observe that each model allocates its distributional judgments in different ranges. For example, \texttt{GPT-4} allocates probabilities in a wide range $[0.0, 1.0]$, while \texttt{Ministral-8B-Inst} allocates them in a much narrower range $[0.19, 0.58]$. However, this variation across models do not represent the varying confidence of the models' interpretive judgments (\Cref{sec:distribution-not-confidence}). Rather than basing analysis on the absolute values of the probabilities, we look at the distributions separately for each model to determine whether a given model provides a useful signal for interpretation.
\paragraph{Instruction tuning is associated with a wider range of judgment probabilities but also introduces unpredictable bias.}\label{sec:instruction-tuning-impact}
Across the board, instruction-tuned models utilize a wider range of judgment probabilities than their base counterparts, with instruction-tuned \texttt{OLMo}, \texttt{gemma}, and \texttt{GPT-4} models utilizing the entire space in [0, 1] as shown in \cref{tab:judgment-distributions}.
However, other changes introduced by instruction tuning are less consistent---the magnitude and direction of changes in both categorical and distributional judgment varies by model.
For example, instruction tuning on \texttt{Llama-8B} leads to a predominance of \negjudg{NotCovered} judgments, while instruction tuning of \texttt{Llama-70B} leads to predominance of \posjudg{Covered} judgment, when prompted using the \texttt{Yes/No} question variant.

\subsection{Robustness to question variation}

For a model to be considered reliable at answering interpretive questions, it should be robust to minor variation in how the question is phrased. As described in \cref{sec:methods-robustness}, we measure model responses while varying the phrasing of the question in the prompt and leaving the content unchanged, and analyze how the variation affects the models' categorical and distributional judgments. We report three major findings in this section of our study.

\begin{figure}[t]
    \centering    \includegraphics[width=0.7\linewidth]{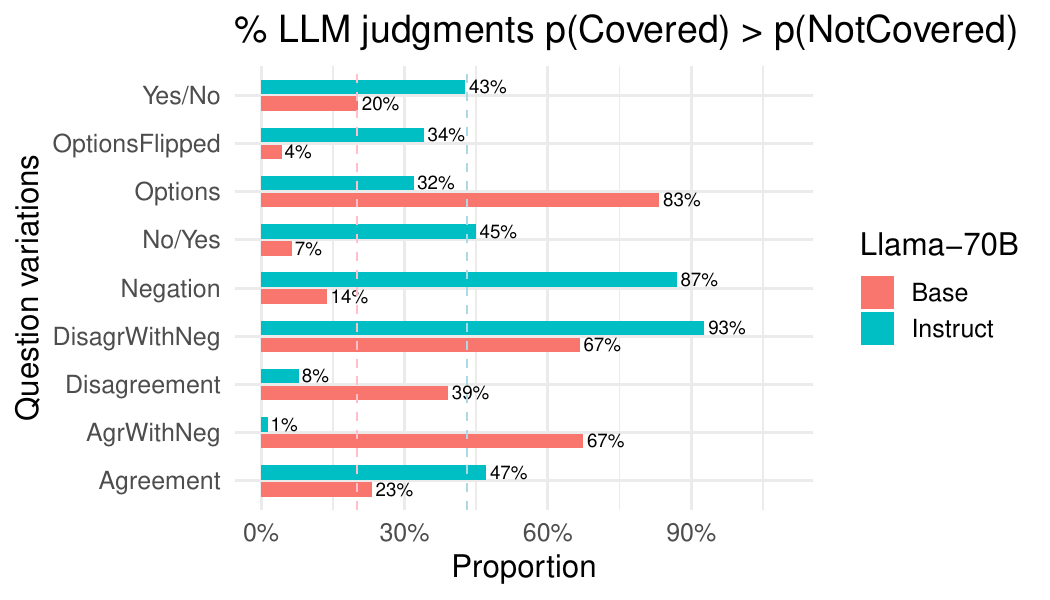}
    \caption{\texttt{Llama-70B} model responses across question variants, each of which results in a large shift away from values in either directions given the \texttt{Yes/No} variant, indicated with the dotted lines.
    }
    \label{fig:prompt-type-bar}
    \Description{Horizonal bar plots of rate of covered LLM judgments from Llama-70B Base and Instruct models, highlighting changes by question variants. They show shifts that are increases and decreases. Some shifts are large such as up to twice, and less than half.}
\end{figure}

Given the nine question variants, for each item and model, one of the categorical judgments (\posjudg{Covered}, \negjudg{NotCovered}) will be the majority judgment and the other the minority judgment. The strength of the majority can vary---from 5 of 9 variants (an indication of brittleness) to 9 of 9 variants (unanimity, indicating robustness).  A frequency table for majority judgments collated by items for each model is shown in \cref{tab:prompt-variation-majority-vote}.

\paragraph{Ubiquitous lack of consistency across question variants.}
\label{sec:prompt-shopping}
We observe lack of consistency across the 9 question variants in each LLM we study.
As illustrated in \cref{tab:prompt-variation-majority-vote}, in 2,061 of 2,070 model item-model combinations (138 scenarios for each of the 15 models), both judgments can be found across the question variants with each model; only in 9 item-model combinations is the judgment fully consistent across all question variants.

Model judgments are sensitive even to variations such as reversing the order of the provided answer choices or introducing a negation word, as seen in  \cref{fig:prompt-type-bar}
for \texttt{Llama-70B} models, where the swap from \texttt{Agreement} to \texttt{AgrWithNegation} prompt type produces a 44\% absolute shift in the base model, and a 46\% absolute shift in the instruction-tuned model, with respect to the rate of categorical \posjudg{Covered} judgments.
This ubiquitous lack of consistency exposes a generalization failure on the part of the models. Worse, it invites users invested in a particular outcome to engage in ``prompt shopping,'' varying the prompt until the desired response is produced \citep{waldon_llms,pruss-25}. 

\paragraph{Some questions are more likely to elicit minority judgments.} 
\label{sec:question-variations-appendix}
While unanimity is difficult to come by, some question variants are still more likely to induce the `minority judgment' which disagrees with the majority of judgments produced by the same model given the same scenario. As shown in  \cref{tab:minority-frequency}, we find that the \texttt{Disagreement} variant yields the minority judgment most frequently, while the \texttt{Yes/No} variant is least likely to yield the minority judgment.
Of nine question variants we investigate, \texttt{Disagreement}, \texttt{AgrWithNegation},  \texttt{DisagrWithNegation} and \texttt{Options} are the most frequent elicitors of minority judgment, accounting for $67\%$ of minority responses.
However, even in the absence of these four variants, unanimity occurs only in $33\%$ of items.

\begin{table}[t!]
    \centering\small
\begin{tabular}{lrrrrr|lrrrrr}
\toprule
\textbf{Model} & \textbf{5} & \textbf{6} & \textbf{7} & \textbf{8} & \textbf{9} & \textbf{Model} & \textbf{5} & \textbf{6} & \textbf{7} & \textbf{8} & \textbf{9} \\
\midrule
Llama-70B   & 33  & 68 & 36 & 1  & 0 & Ministral-8B-Inst & 24 & 65 & 30 & 18 & 1 \\
\quad +Inst & 25  & 33 & 78 & 2  & 0 & OLMo-2-7B         & 46 & 65 & 20 & 7  & 0 \\
Llama-8B    & 40  & 52 & 46 & 0  & 0 & \quad +Inst       & 51 & 57 & 30 & 0  & 0 \\
\quad +Inst & 6   & 39 & 59 & 31 & 3 & Gemma-7b          & 44 & 58 & 29 & 7  & 0 \\
Llama-3B    & 95  & 40 & 3  & 0  & 0 & \quad +it         & 9  & 31 & 79 & 19 & 0 \\
\quad +Inst & 75  & 48 & 15 & 0  & 0 & GPT-4             & 4  & 9  & 57 & 63 & 5 \\
Llama-1B    & 12  & 57 & 69 & 0  & 0 & GPT-2-medium      & 50 & 83 & 5  & 0  & 0 \\
\quad +Inst & 129 & 9  & 0  & 0  & 0 &                   &    &    &    &    &   \\
\bottomrule
\end{tabular}
\caption{Number of items by number of question variants that yielded the majority judgment for the model. For example, there were 33 items for which \texttt{Llama-70B} produced one judgment for 5 variants, and the other judgment for 4 variants. Each judgment is a binary choice between \posjudg{Covered} and \negjudg{NotCovered}.
9 variants producing the same judgment indicates unanimity, which occurs in 3 items for \texttt{Llama-70B-Inst} and in 5 for \texttt{GPT-4}.
}
\label{tab:prompt-variation-majority-vote}
\end{table}

\begin{table}[!ht]
    \centering\small
\begin{tabular}{lrr|lrr}
\toprule
\textbf{Variant} & \textbf{Count} & \textbf{\%Minor.} & \textbf{Variant} & \textbf{Count} & \textbf{\%Minor.} \\
\midrule
Disagr.         & 1256 & 21 & Options F. & 493  & 8 \\
Agr. w/ Neg.    & 1045 & 17 & Negation   & 489  & 8 \\
Disagr. w/ Neg. & 918  & 15 & N/Y        & 275  & 5 \\
Options         & 809  & 14 & Y/N        & 188  & 3 \\
\cmidrule(lr){4-6}
Agr.            & 501  & 8 & \textbf{Total} & \textbf{4975} & \\
\bottomrule
\end{tabular}
\caption{The number of minority judgments for each question variant, and the percentage proportion in minority judgments. The counts are sorted vertically in descending order. An equal proportion would lead to a $0.09$ proportion for each variant.}
\label{tab:minority-frequency}
\end{table}


\paragraph{Some question variants have a stronger distributional effect than others.}
We quantify the impact of each question variant (relative to the default \texttt{Yes/No} variant) by measuring the distance between 1) the judgment distribution given that variant and 2) the judgment distribution given the default \texttt{Yes/No} variant.
\label{ref:JSD}
We measure distribution distance with the Jensen-Shannon distance \cite[JSD; ][]{jsd_2003}, which is based on KL divergence \cite{kullbackInformationSufficiency1951} but is symmetric and provides a distance measurement between $0.0$ (identical distributions) and $1.0$ (maximally different).
The question variant that yields the most distant distribution for each model is listed in \cref{tab:model-prompt-type-jsd}~(\cref{sec:appendix-dist-robustness}), where average distance is obtained by calculating JSD at the item level then aggregating by question variant for each model. The  \texttt{Disagreement with negation} variant most frequently (5 of the 15 models) leads to the biggest change in the distributional judgments in the most models, including up to $0.56$ for GPT-4. The lack of robustness supports our first hypothesis, and should be considered a limitation of LLMs for use in legal interpretation, since such inconsistency shows models are brittle in the face of superficial prompt variation.

\subsection{Correlation to human judgment}
\label{sec:human-correlation}

Finally, we compare LLMs' responses and distributional judgments to human judgment data from \citet{waldon_vague_contracts_2023} to test our hypothesis that model responses are poorly correlated to human judgments.
In the \citet{waldon_vague_contracts_2023} study, each human participant was asked for a response of \postok{yes}, \negtok{no}, or \textsf{\textcolor{gray}{can't decide}} (\postok{yes} and \negtok{no} correspond to \posjudg{Covered} and \negjudg{NotCovered} judgments, respectively).\footnote{More details on the response collection is provided for reference in \cref{sec:appendix-vague-contracts-questions}).} The dataset contains judgments for all 138~items with total of  judgments from 1346 participants, with an average of 30 judgments per scenario. 

\begin{figure*}
    \small
    \centering
    \includegraphics[width=.9\linewidth,trim=0 0 0 30pt,clip]{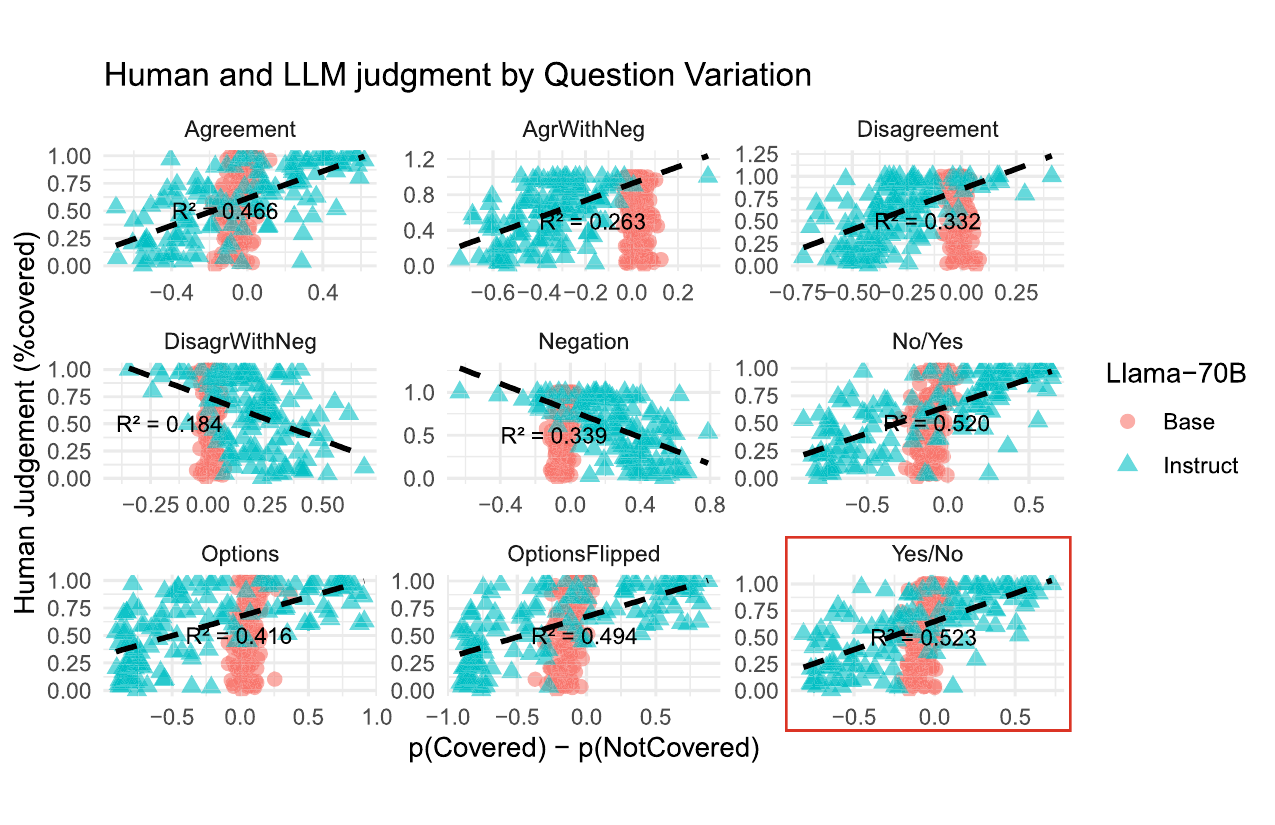}
    \caption{\texttt{Llama-70B} model probabilities versus human consensus across question variants. Dotted lines and the corresponding $R^2$ are best best-fit lines between human and Instruct LLM. The \texttt{Yes/No} question variant, highlighted in red, represents the highest $R^2=0.523$ value for the model, a moderate correlation.} 
    \label{fig:llama-human}
    \Description{Scatterplots of Human judgment variable and LLM judgment variables from Llama-70B Base and Instruct models visualizing the correlation analysis. Each scatter plot visualizes values for one specific question variant but both LLMs, and also shows the best-fit line for Instruct LLM with the corresponding $R^2$ value. The best correlation of $R^2 = 0.523$ is for the Yes/No question variant.}
\end{figure*}

If LLMs are good models of population-level interpretive consensus among lay human speakers, we expect there to be model-internal quantity (or a derived value) to bear a monotonic relationship to the degree of human consensus. In our correlation analysis, the dependent variable is the proportion of human \posjudg{Covered} responses on the same item. We use LLM judgment (a measure derived from LLM token probability) as the independent variable. \label{point:calibration} We explicitly avoid conflating LLM probability with the confidence of the prediction being correct; token probability distributions represent generation probabilities, not estimates of confidence~\citep{pmlr-v267-huang25w}, and current LLMs are not sufficiently calibrated to assume a correlation between probability and confidence~\citep{Ch11CalibrationJurafskyMartin2026}.

Our linking hypothesis posits that LLM judgment can be linearly correlated to human consensus.
We link the endpoints of the LLM-derived measure $p(\posjudg{Covered}) = \{0, 1\}$ to unanimous human judgments and consider four operationalizations that satisfy this boundary condition. We find $R^2$ values to be stable across the four operationalizations (reported in the appendix \cref{tab:operationalizations} for completeness), suggesting that our linking hypothesis is robust to minor variations in formulation. \footnote{We acknowledge that there may be other monotonic linking functions that could generally lead to higher $R^2$, but given the goal and scope of this work, we stick to a linear link as a reasonable default.} The operationalization of LLM judgment used in the rest of the paper is the difference in model probabilities $\greendelta = p( \posjudg{Covered} ) - p(\negjudg{NotCovered})$ due to its benefit in representing consensus in either direction. With this operationalization, our linear linking hypothesis accounts for the probability judgments for both \posjudg{Covered} and \negjudg{NotCovered} with the decision boundary at 0. A perfect correlation would be fit to the equation \%\posjudg{Covered} $= 0.5\greendelta + 0.5$.

We use this linking function to perform a $R^2$ analysis \citep{wright1921correlation} with LLM judgment as the independent variable in determining the proportion of human majority judgment. Our analysis is not a benchmarking objective, rather an analytical view at how much signal of \textit{human} judgment they provide.

\paragraph{Only some LLMs are moderately correlated to human judgment some of the time.} 
Our results show that only larger instruction-tuned models' judgments exhibit some correlation to human judgment, and only for some specific question variants. This is despite evidence for scaling--models do show better fit to human data and stronger performance in predicting human consensus as the number of parameters (and likely the amount of training data) increases.
In particular, only instruction-tuned models with 70 billion parameters or more report an $R^2$ value greater than $0.5$, as shown in \cref{tab:strongest-correlation-by-model} and \cref{fig:best_prompt_for_each_model}. 
Correlation between \texttt{Llama-70B-Inst} response and human judgment across question variants is illustrated in \cref{fig:llama-human}, where one variant yields a negative correlation ($m < 0$) with a $R^2$ value of 0.18.
The significant variation in $R^2$ values across question variants suggests that the models' responses are not, in general, representative of human-like judgment, but rather highly influenced by prompt design and question form.
This highlights a second limitation to LLMs in legal interpretation: at best, the models we examine achieve moderate correlation to human judgment in select model-question variant pairs.

Furthermore, even when there is correlation between token probabilities and human judgment, the output tokens can be overwhelmingly biased. 
\texttt{Llama-70B-Inst}'s responses to the \texttt{AgrWithNeg} variant shown in \cref{fig:llama-human} illustrates that despite a (weak) correlation of \mbox{$R^2=0.26$} when examining gradations of probabilities, the actual responses are overwhelmingly \negjudg{NotCovered}, where $\greendelta < 0$, further raising caution to LLMs' use for legal interpretation.

\begin{table*}[h!]
    \centering
    \small
\begin{tabular}{@{}llrr|rr|llrr|rr@{}}
\toprule
\multicolumn{2}{c}{ }&\multicolumn{2}{c}{\textbf{Corr. Analysis}}&\multicolumn{2}{c|}{\textbf{Oracle }}& \multicolumn{2}{c}{ }&\multicolumn{2}{c}{\textbf{Corr. Analysis}}&\multicolumn{2}{c}{\textbf{Oracle}}\\
\textbf{Model} & \textbf{Var.} & $\bm{m}$ & $\bm{R^2}$ & \textbf{Acc.} & \textbf{Thr.} & \textbf{Model} & \textbf{Var.} & $\bm{m}$ & $\bm{R^2}$ & \textbf{Acc.} & \textbf{Thr.} \\
\midrule
Llama-70B & Opt. F. & 2.0 & 0.25 & 0.69 & $-$0.17 & Ministral-8B-Inst & Yes/No & 0.97 & 0.27 & 0.69 & $-$0.11 \\
\quad +Inst & Yes/No & 0.65 & \textbf{0.52} & \textbf{0.79} & $-$0.28 & OLMo-2-7B & No/Yes & 1.3 & 0.07 & 0.64 & $-$0.10 \\
Llama-8B & Agr. & 1.1 & 0.04 & 0.58 & $-$0.02 & \quad +Inst & Yes/No & 0.21 & 0.18 & 0.64 & $-$0.63 \\
\quad +Inst & Options & 1.1 & 0.20 & 0.67 & $-$0.36 & gemma-7b & Yes/No & 1.4 & 0.10 & 0.59 & $-$0.10 \\
Llama-3B & Disagr. & $-$0.67 & 0.03 & 0.42 & 0.11 & \quad +it & Neg. & 0.21 & 0.06 & 0.60 & $-$0.16 \\
\quad +Inst & Yes/No & 0.46 & 0.06 & 0.59 & $-$0.25 & GPT-4 & Yes/No & 0.26 & \textbf{0.60} & \textbf{0.83} & $-$0.25 \\
Llama-1B$^*$ & No/Yes & $-$0.91 & 0.01 & 0.38 & 0.13 & GPT-2-medium$^*$ & Options & 0.43 & 0.01 & 0.62 & $-$0.19 \\
\cmidrule(lr){7-12}
\quad +Inst$^*$ & Neg. & $-$0.37 & 0.01 & 0.38 & 0.34 & \multicolumn{2}{c}{Majority class baseline} & & & 0.61 \\
    \bottomrule
\end{tabular}
\caption{
For each model, we report the question variant with the highest correlation to human judgments along with the $R^2$. We also report the accuracy with an oracle threshold (knowing the labels) of a binary classifier predicting the human majority judgment from the LLM probability difference $\greendelta$.
We offer the always-\posjudg{Covered} accuracy of the majority class as a reference baseline.
The correlations except three indicated with a $^*$ have a p-value $< 0.05$. These are also the three weakest correlations.
Unlike p-values reported for performance with random sampling effects, the p-values provided is a significance measure of the~\textit{best-fit} line $m$ and $b$, rather than the coefficient of determination $R^2$.
Refer to \cref{fig:best_prompt_for_each_model} for a detailed visualization of each model-prompt pair.
An oracle model uses LLM judgment to linearly predict binary human consensus.}
\label{tab:strongest-correlation-by-model}
\vspace{-10pt}
\end{table*}

\paragraph{Even the best-correlated LLMs are unreliable predictors of human judgment.}
As another measure of correlation, we consider an oracle threshold classifier that predicts the binary human judgment as a function of the LLM's probability-difference $\greendelta$.
The threshold is set based on the human data to match 
where the best-fit line crosses $0.5$ in human judgment.
Given a probability difference value exceeding the threshold, the classifier predicts the \posjudg{Covered} judgment; otherwise, it predicts \negjudg{NotCovered}.
 Intuitively, this tells us how accurate the LLM would be at giving probabilities consistent with the human majority judgment, if only we knew how to perfectly interpret the model-specific probability scale. Binary classification accuracies appear in the penultimate column of \cref{tab:strongest-correlation-by-model}.

The model-variant pair best correlated to human judgment, \texttt{GPT-4} responses to \texttt{Y/N} ($R^2=0.60$), yields a linear model that accurately predicts human consensus 83\% of the time.
Even if this may initially appear satisfactory, we contend that even 1 in 6 binary classification error rate in an interpretative tool as considered in \citet{hoffman_generative_interpretation_2024, snellvsunited, unitedstatesvsdeleon} defeats its purpose.
For comparison, a majority-class classifier always voting \posjudg{Covered} would be correct 61\% of the time.

We describe an example item where \texttt{GPT-4} judgment drastically differs from human judgment in \cref{fig:gpt-4-wrong-example}, where \texttt{GPT-4} overwhelmingly prefers the \posjudg{Covered} judgment to the \negjudg{NotCovered} judgment with $p($\posjudg{Covered}$) - p($\negjudg{NotCovered}$)=0.85$.
For comparison, only $17\%$ of human judgments predicted \posjudg{Covered}, preferring the \negjudg{NotCovered} judgment.

\begin{figure}[!ht]
\small
    \begin{tcolorbox}\sf
Belinda's home insurance policy includes coverage for ``Escape of Water'' damage, defined as ``damage to buildings and contents caused by a sudden, unforeseen flow of water from an internal source, such as a burst pipe or faulty appliance.''

        \begin{tcolorbox}
            Belinda lives in a two-story house with a bathtub on the second floor. One day, Belinda, a sculpture artist, fills the tub with paint in order to dip her sculptures in paint, but the tub suddenly cracks and begins to leak, causing the paint to seep through her floor and damage her walls. Belinda files a claim with her insurance company for the damage.
        \end{tcolorbox}
        Considering just how ``internal source'' would be understood by ordinary speakers of English, is Belinda covered by the insurance---yes or no? Final answer:
    \end{tcolorbox}
    \caption{Prompt to which GPT-4 overwhelmingly prefers the \posjudg{Covered} judgment to the \negjudg{NotCovered} judgment with $\Delta=0.85$, compared to human judgment \%\posjudg{Covered} $= 0.17$.}
    \label{fig:gpt-4-wrong-example}
    \Description{Figure 4. Fully described in text.}
    \vspace{-10pt}
\end{figure}

\paragraph{Response incongruence: answering the wrong question.}
On some question variants, model responses are sometimes negatively correlated to human judgment when the question includes negation, which suggests that LLMs' attested difficulties with negation \citep{garcia-ferrero-etal-2023-dataset, varshney-etal-2025-investigating} persist in legal interpretation.
However, we find this behavior to be unpredictable across model type, question variant, and scenario. We observe negative correlation $m < 0$ between human judgment and difference in LLM judgment probability in \texttt{DisagrWithNeg} and \texttt{Negation} prompt types in \cref{fig:llama-human}, in \texttt{DisagrWithNeg} prompt type in \cref{fig:gpt-human}. Another version of response incongruence is due to failure in instruction-following. From a manual review of all 1,242 responses from \texttt{Llama-70B}, we find that a total of 14 responses to \texttt{Options} and \texttt{OptionsFlipped} may be incongruent, conveying judgment in explicit response rather than selecting a provided choice.
See \Cref{sec:token-output-alignment} ~for more detail.

On the most optimistic interpretation of the results---that the largest models with certain prompt types achieve nontrivial correlation, which might be improved with further engineering to the point of reliability---we underscore two points of caution: (a)~we have not ruled out the possibility of data contamination (\Cref{sec:appendix-contamination}), and (b)~our experimental data is specific to insurance contract scenarios. Thus, there is a long road ahead for any efforts to develop and verify LLMs for practical use.

\section{Related Work}
\label{sec:related-work}

Legal NLP, legal language, and LLM robustness are popular areas of research. Nonetheless, there is limited scholarly work on the utility of LLMs for novel interpretation questions in the law, and most of it is based on conceptual arguments and case studies rather than controlled experimentation.

In this section, we engage with scholarship on LLM-based legal interpretation, as well as some broadly related studies in legal NLP.

\paragraph{Legal interpretation.}
Our study asks whether LLMs can reliably answer novel questions about ordinary meaning.
Several recent papers have expressed skepticism.
\Citet{leeArtificialMeaning2024} contrast LLMs with corpus linguistics, arguing that LLM interpretations are less replicable, transparent, and generalizable.
\Citet{waldon_llms}  argue that the way LLMs are designed makes them vulnerable to misinterpretation and misuse. 
In a similar vein, \citet{grimmelmann2025generative} argue that LLMs are unproven for legal interpretation due to a \textit{reliability} gap (LLMs are not always consistent and reproducible), and an \textit{epistemic} gap (interpretive conclusions in text produced by the model are not necessarily accurate measurements of ordinary meaning as understood by humans).
Our robustness and correlation evaluations are ways of quantifying the respective gaps.

In prior literature, claims about LLM reliability have been supported with ad~hoc case studies---with the exception of ~\citet{choiOfftheShelfLargeLanguage2025}, who conducted a set of experiments to test LLMs' sensitivity to prompt variation.
While our goal and conclusions are similar to those of \citet{choiOfftheShelfLargeLanguage2025}, there are important methodological differences:
(\romannumeral 1)~\citeauthor{choiOfftheShelfLargeLanguage2025} utilized 5 contract scenarios from \citet{hoffman_generative_interpretation_2024} while we use 138 insurance policy scenarios \citep[from][]{waldon_vague_contracts_2023}; and (\romannumeral 2)~we perform controlled, systematic variation of the questions for our studies rather than utilizing LLMs to generate prompt variants. Both our results and \citeauthor{choiOfftheShelfLargeLanguage2025}'s underscore the need for caution when contemplating applying LLMs to legal interpretation. 

Other studies have compared LLMs' interpretive judgments to those of lay human respondents. \Citet{waldon_vague_contracts_2023} find that InstructGPT \cite{ouyangInstructGPT} over-predicts human consensus in their interpretive scenarios but that few-shot prompting mitigates this behavior. With the same items and human judgments, we conduct a different set of experiments focused on analyzing a range of newer models. In a follow-up to our study, \citet{petersen-26} collect new human judgments on materials based on \citeauthor{waldon_vague_contracts_2023}'s stimuli and investigate susceptibility to prompt variation in a newer model (GPT-5 with and without reasoning). \citeauthor{petersen-26} find that while GPT-5 with `reasoning' is relatively robust to prompt variation, correlation to human judgment is still lacking.

\citet{martinezComputationalCanons2025}, using a different dataset derived from previously adjudicated U.S.\ cases that involved ordinary (or plain) meaning analysis\footnote{We refer readers to \citet{basile2024ordinary} for further discussion of these concepts.}, compares interpretive judgments of different human groups (judges, lawyers, and laypeople) and LLMs with \YesNo questions. Most notably in the context of our study, \citet[p.~49]{martinezComputationalCanons2025} reports that \texttt{GPT-4}'s response is consistent with the majority of laypeople for 78\% of the items. (Filtering to the subset of items with a clear majority among laypeople, the model is consistent with that majority 83\% of the time.)
\citet[p.~63]{martinezComputationalCanons2025} concludes that, when compared against the \emph{mismatch} of judgments between judges, lawyers, and layman surveyed, this shows ``LLMs can, under controlled conditions, reliably track the consensus \dots at at least as high a rate as the court''. 
Quantitatively, our results with different data and experimental methodology are broadly compatible: note the 83\% oracle accuracy for \texttt{GPT-4} in \cref{tab:strongest-correlation-by-model}.
Nevertheless, we take a different perspective on the implications.
We argue that, even if prompt instability were not a problem, judges should think twice before taking at face value a system that gives the wrong answer a fifth of the time (without a well-calibrated indication of confidence).

\paragraph{Scope of other studies on LLM instability in legal interpretation}
While this paper is not the first to report that LLMs are unstable when it comes to legal interpretation, prior studies differ with respect to their empirical scope.

\Citet{waldon_llms} perform a red-teaming exercise grounded in the `directed queries' employed in \citet{snellvsunited} and \citet{unitedstatesvsdeleon}. They then perform subtle variations to the queries specifically to elicit counter (metalinguistic) judgments. This demonstrates instability in legal interpretation with an adversarial approach. Their investigation utilizes two samples and specific manipulations for each to elicit counter judgments.

\citet{choiOfftheShelfLargeLanguage2025} is grounded in the five scenarios used in \citet{hoffman_generative_interpretation_2024}. \citeauthor{choiOfftheShelfLargeLanguage2025} utilizes Claude 3.5 \cite{Claude3S} to generate paraphrases of a scenario, including the question and the term under interpretation. \citeauthor{choiOfftheShelfLargeLanguage2025} generates 2000 paraphrased scenarios for each of the five scenarios, and utilize a relative probability measure (between the two options) to look at instability. 

By contrast, our investigation utilizes 138 human-constructed scenarios based on 46 real-world policy scenarios from \citet{waldon_vague_contracts_2023}. Additionally, our variations are designed hold the literal content of the question presented as constant as possible while \citeauthor{choiOfftheShelfLargeLanguage2025}'s LLM-generated variations at times involve more substantial changes in wording. Moreover, our prompt variants are systematic, without adversarial manipulations, and do not use model-generated data. (\Citet{petersen-26} echo our design in this respect but employ fewer variants.)

\paragraph{Legal NLP} In the broader space of Legal NLP~\cite{NLP4Legal2025Survey}, two recent studies are especially relevant to this work. \Citet{luo-etal-2025-automating} instruct LLMs to consider a target legal term (e.g. `landscaping') and generate, based on legal documents from previous cases, an interpretive explanation (along with conditions governing the applicability of the interpretation). 
This task concerns the extraction of a previously articulated interpretation, rather than novel conclusions about meaning.
They report system performance comparable to that of human experts for the task, but they do not evaluate models for robustness across prompts.

\Citet{unstable_answers_2026} investigate LLM instability in the context of a legal judgment prediction task, where the models are prompted with a synopsis of the case and asked to predict which party should prevail. 
This task demands full accounting of all the issues in a particular case, rather than a focused interpretive question. There is no indication that the cases were selected with emphasis on language interpretation. 
Investigating closed commercial models with a `reasoning' step, they 
find significant instability; i.e., the answer is not even consistent given the same model and prompt.

\section{Conclusion}
Given the excitement about and increasing critical legal scholarship on the use of LLMs for legal interpretation, we conducted a systematic study of this capability with a focus on LLM judgments regarding the `ordinary meaning' of legal language, formulated as binary-choice QA. We examined LLMs' judgments for both robustness to prompt variation and correlation to human judgments, finding:
(i)~Some LLMs behave like stopped clocks, with a strong tendency to provide the same judgment for most input, regardless of the scenario.
(ii)~Models show a ubiquitous lack of consistency across question variants.
(iii)~Correlation with human judgment is at best moderate, and is strongest in larger, instruction-tuned models.

Our findings inform a growing body of scholarship on the suitability of LLMs for legal interpretation. The experiments establish an informed evaluation method along two axes: robustness and human correlation. In addition to raising practical concerns regarding LLM legal interpretation, our work addresses some of the theoretical considerations at the heart of the debate. LLM development continues to innovate across data, architecture, training, and inference, keeping pace with an ever-growing range of uses. However, for at least some proponents, the excitement around LLMs for legal interpretation is not due to the most recent innovations in LLM system design but rather to the assumption that large-scale data (seen in pretraining) endows the model with sophisticated language ability and the capacity to reflect on general patterns of English usage. Our experiments focused on models with limited post-training (except for GPT-4) cast doubt on this assumption. Both base and instruction-tuned models struggle to reach satisfying levels of robustness and accuracy when it comes to ordinary meaning analysis. Our results thus suggest that large-scale pretraining is not enough to produce models that are useful for the task at hand.

Our results furthermore show that LLMs are strongly affected by training choices (e.g. instruction tuning significantly shifts judgment probability distribution) and size (e.g. degree of correlation increases with model size). Even an extensively engineered model (GPT-4) that has undergone pretraining at commercial scale can fail to be robust and show limited correlation to human judgments.

What does this mean for the use of LLM legal interpretation on the bench? Our results and findings strongly caution against Judge Newsom's approach of directly asking an LLM to settle a question of `ordinary' linguistic meaning. Although ChatGPT and other widely available systems boast fluent and easy-to-query chatbots, we cannot assume that their underlying LLMs give sound answers to such questions.
This is not to say, however, that judges could not use LLMs productively in other ways for language interpretation, especially in concert with other empirical or analytical tools (e.g., \citet{waldon_llms} offer a proposal for `dialectical AI' that does not rely on assuming the system has good judgment). More nuanced approaches should be the focus of future work. 

Our experimental coverage of LLMs is, of course, not exhaustive; perhaps a newer model or approach (e.g., involving a bespoke system or user-facing design) will mitigate some of these limitations.
Any such suitability for this task should be demonstrated in rigorous experiments beyond those laid out in this paper. 

But the evidence thus far spectacularly fails to meet the burden of proof.

\begin{acks}
This research was supported in part by NSF award IIS-2144881.
The experimental portion of this work was made possible by the Georgetown University High Performance Computing Cluster, Calcul Québec, and the Digital Research Alliance of Canada.
We thank Kevin Tobia, Ethan Wilcox, and Amir Zeldes, Wisdom Obinna, Mohammed Ahmed, members of the NERT lab, anonymous reviewers, and attendees of SOLID (Symposium on Legal Interpretation and Data) 2026 for helpful discussions and feedback that informed this work. 
\end{acks}

\section*{Generative AI usage statement}

We utilized ChatGPT \cite{achiam2023gpt}, \href{https://github.com/features/copilot}{GitHub Copilot}, and \href{https://www.jetbrains.com/pycharm/features/ai/}{AI Assistant in Pycharm} during the implementation of our experiments.

\bibliographystyle{ACM-Reference-Format}
\bibliography{custom}

\appendix
\section{Prevelance of ordinary and plain meaning in US Judicial System}
\label{sec:appendix-other-cases}
To highlight the prevalence and importance of ordinary (and plain) meaning analysis, we provide quick reference to some well known and important cases across the legal spectrum below.

\begin{itemize}
\item \textbf{Criminal law:} The term `physically restrained' in a sentencing provision was interpreted in \textit{U.S.~v.~Deleon} \citep{unitedstatesvsdeleon}, with one of the judges exploring the use of AI (as noted above).
\item \textbf{Civil rights law:} The phrase `because of sex' in a federal statute was interpreted to protect against discrimination on the basis of sexual orientation and gender identity in the U.S.~Supreme Court case \textit{Bostock v. Clayton County} \citep{bostock-vs-claytoncounty}.
\item \textbf{Regulatory law:} The U.S.~Supreme Court upheld an agency's interpretation of the term `firearm' as encompassing weapons parts kits in \textit{Bondi v.~VanDerStok} \citep{bondi-vs-vanderstok} (see also \citep{waldon2024reading}).
\item \textbf{Constitutional law:} 
\textit{District of Columbia v Heller} \citep{district-vs-heller} concerning 2nd amendment rights,
\item \textbf{Immigration law:}
\textit{Campos-Chaves v.~Garland} \citep{campos-chaves-vs-garland} concerned the circumstances under which a judge could rescind an order to deport a noncitizen.
Parties in the case disagreed about the interpretation of a condition that was formulated as a negative disjunction (\textit{not A or B}).
\end{itemize}

\section{Additional details on experimental materials}
\label{sec:appendix-vague-contracts}

\citet{waldon_vague_contracts_2023} collected human judgments for interpretive questions about insurance contract vignettes. Here we detail the structure and topics of these vignettes.
\Cref{fig:vague-contracts-vignette} illustrates the kind of text seen by human subjects in their study:

\begin{figure}[!ht]
\small
    \begin{tcolorbox}\sf
Steve's car insurance policy includes coverage for ``Vehicle Damage,'' defined as ``loss or damage to the policy holder's 1) car; or 2) car accessories (while in or on the car).''

        \begin{tcolorbox}
            One day, Steve is involved in a minor accident. His GPS navigation system, which was in the car at the time, was damaged. Steve files a claim with his insurance company for the damage.
        \end{tcolorbox}
        \begin{enumerate}
            \item Do you think that the claim is covered under Vehicle Damage as it appears in the policy? [\posjudg{Yes} / \negjudg{No} / \textsc{Can’t Decide}] \item You are one of 100 people who have volunteered to answer these questions. How many of the 100 do you think will agree with your answer to question (1)? \item How confident are you in your answer to question (1)? [(Not at all / Slightly / Moderately / Very / Totally) confident]
        \end{enumerate}
    \end{tcolorbox}
    \caption{An example vignette from the questionnaire provided to the participants by \citet{waldon_vague_contracts_2023}. The vignette corresponds to one of the 138 items. Since our study focuses on interpretative judgment, question 1 is of interest to us, and responses make up the human judgments used in correlation analysis in \Cref{sec:human-correlation}. }
    \label{fig:vague-contracts-vignette}
    \Description{Figure 5. Fully described in text.}
\end{figure}

Each vignette consists of a hypothetical contract provision with a term and definition, coupled with a hypothetical scenario meant to test the interpretation of that definition (\cref{tab:prompt-structure}). The provisions are drawn from a range of insurance types (\cref{tab:contract-items}). The term at issue, or \emph{locus of uncertainty}, depends on the scenario; the full list of these terms appears in \cref{tab:contracts-locus}.
We note that these vignettes were artificially constructed, but are meant to imitate real-world insurance contract scenarios.


\begin{table}[!ht]
    \centering\small
    \begin{tabular}{lll}
    \toprule
    \multicolumn{3}{c}{\textbf{Insurance Types}}\\
    \midrule
    Emergency Damages & Hot Work & Public Liability Property Damages \\
    Escape of Oil & House Removal & Storm Damage \\
    Escape of Water & Identity Theft & Trace and Access \\
    Fire & Loss and Accidental Damage & Vehicle Damage \\
    Flooding & Loss or Damage to a Goods Carrying Vehicle & Vehicle Fire \\
    Garden Plants & Malicious Acts or Vandalism & Vehicle Glass \\
    General Damages & Personal Accident & Vehicle Theft \\
    Ground Heave & Personal Accidents & Wind Damage \\
    Hail Damage & & \\
    \bottomrule
    \end{tabular}
    \caption{The unique insurance types represented in \citeauthor{waldon_vague_contracts_2023}'s experimental materials.}
    \label{tab:contract-items}
\end{table}

\begin{table}[hbt!]
    \centering\small
    \begin{tabular}{lll}
    \toprule
    \multicolumn{3}{c}{\textbf{Loci of Uncertainty}} \\
    \midrule
    accessory                 & flammable or combustible materials & permanent or total loss \\
    accidental                & flow of water                      & political disturbance \\
    audio equipment           & glass                              & professional movers \\
    broken glass              & ground heave                       & rapid build-up \\
    causative ``from''        & hard surface                       & reasonable steps \\
    cause                     & heating installation               & regular working conditions \\
    connected with business   & internal source                    & requires / uses / produces \\
    connected with occupation & key theft                          & sudden/unforeseen \\
    custody or control        & leaking                            & temporarily removed \\
    damage                    & malicious people or vandals         & third party \\
    deliberate                & naturally occurring fire            & tracking device \\
    family or employee        & necessary and reasonable           & traveling in \\
    fire damage               & outside the building               & wear and tear \\
    first responder           & perceived emergency                & \\
    \bottomrule
    \end{tabular}
    \caption{Loci of uncertainty in contract provisions. (Note that these are descriptions in the dataset; the actual text prompts in \citet{waldon_vague_contracts_2023} did not contain metalinguistic terms like \emph{causative}.)}
    \label{tab:contracts-locus}
\end{table}

\paragraph{Human Data}
\label{sec:appendix-vague-contracts-questions}
In \citet{waldon_vague_contracts_2023}, the subjects were asked three questions (see \cref{fig:vague-contracts-vignette}), first asking for judgment, second asking for confidence, third asking for a estimation of consensus. For this work, the first and second question would be of interest. However, the second question collects confidence on a discrete likert scale, and hence we do not use it to scale the discrete judgment provided by the subjects.

\section{Data contamination}
\label{sec:appendix-contamination}
Since our source data is from \citeyear{waldon_vague_contracts_2023}, it is possible that it was part of the training for one or more models in our selected suite of LLMs, especially the larger models which show the best correlation.
\Cref{tab:training-cutoffs} catalogs the known cutoff dates for for the models. Any model with the exception of GPT-2 may have been partially or wholly exposed to the dataset.

\begin{table*}[ht!]
    \centering
    \small
    \begin{tabular}{lll}
    \toprule
         \textbf{Model} &  \textbf{Reported 
         Cutoff Date} & \textbf{Data available before cutoff}? \\
         \midrule
         GPT-4 & Dec 1, 2023 & Yes\\
        GPT-2 & Unknown (Released November 2019) & No\\
         Llama-3 & December 2023 & Yes \\
         OLMo-2 & December 2023 & Yes\\
         Ministral-8B & Unknown (Released October 2024) & Maybe\\
         Gemma-3 & August 2024 & Yes\\
       \bottomrule
    \end{tabular}
    \caption{Reported cutoff date for training data or if unknown, model release date for each model. It is possible that our source evaluation data was included in the training dataset for 14 out of 15 models we use for our study.}
    \label{tab:training-cutoffs}
\end{table*}

\section{Alignment between first token probability and decoded output}
\label{sec:appendix-dist-robustness}
\label{sec:token-output-alignment}
In \cref{sec:binary-interp} we described our operationalization of LLM judgment using first output token probability. 
However, this is a simplifying assumption: it is possible that (the probability of) the first token alone misrepresents the full LLM response.
We manually reviewed all outputs of \texttt{Llama-70B} to better understand the effects of our choice to represent LLM judgment by first token probability.
\cref{tab:tokenprob-output-alignment} organizes our analysis. This decision is overall beneficial, allowing us to extract judgment from ambiguous or incoherent responses while extracting an equivalent human-extractable answer in 1,217 out of 1,242 (97.3\%) responses.
Our setup is erroneous in 10 responses (0.8\%), yet enables judgment extraction in 23 non-human-extractable responses due to incoherence or ambiguity (1.9\%).

\begin{table}[ht]
    \centering\small
    \begin{tabular}{lrp{9cm}}
    \toprule
    \textbf{Manual extraction yields} & \textbf{Count} & \textbf{Example} \\
    \midrule
        Correct extraction &  1,217 &  \\
        No extractable answer & 18 & yes or no? Peter has insurance that covers "Loss or Damage to a Goods Carrying Vehicle," which includes "key replacement in the case of theft."
As part of his bakery business, Peter owns a van which he uses to make local deliveries. One day, Peter realizes he has lost the key to the van. \\
        Ambiguous answer & 5 & yes or no? I think the answer is no, because the damage was not sudden. It took several hours for the damage to occur. The damage was not unforeseen, because Carol knew about the leak and did not fix it immediately.
I think the answer is yes, because the damage was sudden. It happened in\\
        Incorrect extraction & 10 & yes or no. The answer is "no." The insurance policy covers damage to belongings that occurs while being stored by professional removal contractors. Bill is not a professional removal contractor, so the damage to Harper's belongings is not covered by the insurance policy \\
        \bottomrule
    \end{tabular}
    \caption{Counts and examples of correct, no, ambiguous, and incorrect extractions in responses generated by \texttt{Llama-70B Base}. Examples are all responses to the \texttt{Yes/No} prompt.}
    \label{tab:tokenprob-output-alignment}
\end{table}

A manual sweep also uncovered other examples of incongruence, specifically in the \texttt{OptionsFlipped} and \texttt{Options} question variants (challenges with extracting multiple choice answers are relatively well studied; \citet{wang2024look, rottger-etal-2024-political}).~In \texttt{Llama-70B} responses, a total of 14 could be considered incongruent, providing answers in plain text rather than selecting one of the \texttt{A.} or \texttt{B.} options.
In 8 of them, first token probability aligns with manually extractable judgment, e.g.~responding with ``Denise is covered. ...'' with $p(\posjudg{Covered}) > p(\negjudg{NotCovered})$.
In the remaining 6, first token probability does not align, e.g.~responding with ``Gavin is not covered'' but also with $p(\posjudg{Covered}) > p(\negjudg{NotCovered})$.

\begin{table}[!ht]
    \centering
    \small
\begin{tabular}{lccc|lccc}
\toprule
\textbf{Model} & \textbf{3} & \textbf{4} & \textbf{5} & \textbf{Model} & \textbf{3} & \textbf{4} & \textbf{5} \\
\midrule
Llama-70B      & 22  & 21  & 95  & Ministral-8B-Inst & 16  & 67  & 55  \\
\quad +Inst      & 24  & 93  & 21  & OLMo-2-7B         & 71  & 26  & 41  \\
Llama-8B       & 33  & 43  & 62  & \quad +Inst    & 36  & 77  & 25  \\
\quad +Inst      & 13  & 98  & 27  & Gemma-7b          & 45  & 50  & 43  \\
Llama-3B       & 11  & 113 & 14  & \quad +it         & 9   & 47  & 82  \\
\quad +Inst      & 57  & 48  & 33  & GPT-2-medium      & 67  & 71  & 0   \\
Llama-1B       & 4   & 25  & 109 & GPT-4             & 5   & 41  & 92  \\
\cmidrule(lr){5-8}
\quad +Inst      & 13  & 125 & 0   & \textbf{All}      & \textbf{426} & \textbf{945} & \textbf{699} \\
\bottomrule
\end{tabular}
\caption{Number of items by number of question variants that yielded the majority judgment for the model. For example, there were 5 items for which GPT-4 produced one judgment for 3 variants, and the opposite judgment for 2 variants. Each judgment is a binary choice between \posjudg{Covered} and \negjudg{NotCovered}. This is a replication of \cref{tab:prompt-variation-majority-vote} but without the four most minority response-inducing variants \texttt{Disagreement}, \texttt{AgreementWithNegation}, \texttt{DisagrWithNegation} and \texttt{Options}.}
\label{tab:prompt-variation-majority-vote-easy-variations}
\end{table}

\begin{table}[!ht]
\centering\small
\begin{tabular}{llrr|llrr}
\toprule
\textbf{Model} & \textbf{Variant} & \textbf{Mean} & \textbf{Std} & \textbf{Model} & \textbf{Variant} & \textbf{Mean} & \textbf{Std} \\
\midrule
Llama-70B  & Options         & 0.09 & 0.04 & OLMo-2-7B         & Disagr. w/ Neg. & 0.37 & 0.05 \\
\quad +Inst  & Negation        & 0.34 & 0.17 & \quad +Inst       & Agr. w/ Neg.    & 0.48 & 0.23 \\
Llama-8B   & Options F.      & 0.15 & 0.04 & Ministral-8B-Inst & Options         & 0.16 & 0.04 \\
\quad +Inst  & Disagr. w/ Neg. & 0.18 & 0.06 & Gemma-7B          & Disagr. w/ Neg. & 0.08 & 0.03 \\
Llama-3B   & Agr. w/ Neg.    & 0.19 & 0.06 & \quad +it         & Options         & 0.78 & 0.04 \\
\quad +Inst  & Options F.      & 0.22 & 0.07 & GPT-2-medium      & Disagr. w/ Neg. & 0.17 & 0.02 \\
Llama-1B   & Options         & 0.28 & 0.05 & GPT-4             & Disagr. w/ Neg. & 0.56 & 0.30 \\
\quad +Inst  & Options F.      & 0.32 & 0.04 &                   &                 &      &      \\
\bottomrule
\end{tabular}
\caption{The question variant for each model with the largest Jensen-Shannon distance from the \texttt{Yes/No} question. Higher distance indicates greater difference between probability distributions.}
\label{tab:model-prompt-type-jsd}
\end{table}


\section{Implementation and Compute}
\label{sec:implementation-details}

\subsection{Load-and-infer pipeline.}
We use \texttt{vllm} \cite{kwon2023efficient} to implement our inference pipeline and use the model implementations available on \url{https://huggingface.co/models}.

All our inference was completed on Tesla L4 GPUs with 24GB of memory, with the exception of \texttt{Llama-70B}, which were run on 4xA100, each with 40GB of memory. Our inference configuration will be available as part of our public code.

\paragraph{Judgment as a sum of probability}
\label{sec:judgment-as-sum}
In \cref{sec:binary-interp}, we consider LLM judgment as a sum of token probabilities that correspond to each judgment. That is, when answering \postok{yes} to the question would indicate the \posjudg{Covered} judgment, we consider $p$(\posjudg{Covered}) = $p$(\postok{Yes}) + $p$(\postok{yes}) + $p$(\postok{YES}). 

\paragraph{Random seed and temperature}
Because a significant portion of our study works with token probabilities, we set temperature to 0, and hence there is no randomness in our inference pipeline.

\subsection{OpenAI GPT-4 Inference with APIs}
\label{sec:openai-appendix}
We used OpenAI API platform for inference with GPT-4. We use \texttt{temperature=0} to get the highest determinism. We use the GPT-4 model with the model identifier \texttt{gpt-4-0613}.

\section{Additional details on LLM response across model, question variants}
\label{sec:additional-details-results}
\begin{figure*}[ht]
    \small
    \centering
    \includegraphics[width=\linewidth]{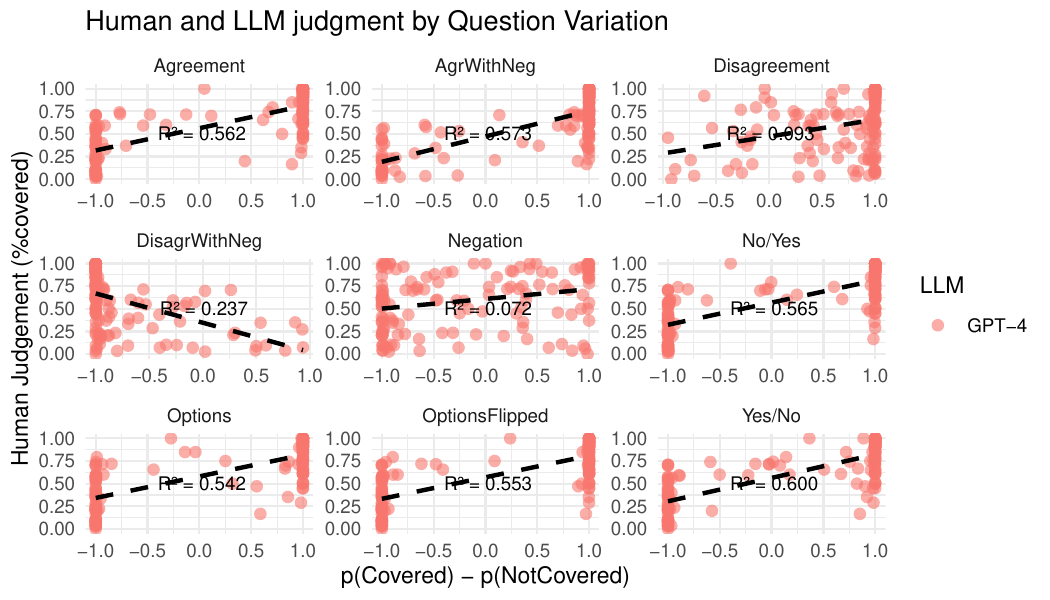}
    \caption{\texttt{GPT-4} judgment probabilities versus human consensus across question variants. Dotted lines are best best-fit lines between human and instruction-tuned LLM.
    } 
    \label{fig:gpt-human}
    \Description{Scatterplots of Human judgment variable and LLM judgment variables from GPT-4 models visualizing the correlation analysis. Each scatter plot visualizes values for one specific question variant, and also shows the best-fit line for Instruct LLM with the corresponding $R^2$ value. The correlations vary from $R^2=0.072$ for the Negation variant to $R^2=0.600$ for the Yes/No variant.}
\end{figure*}

\begin{figure*}[ht]
    \centering
    \includegraphics[width=1\linewidth]{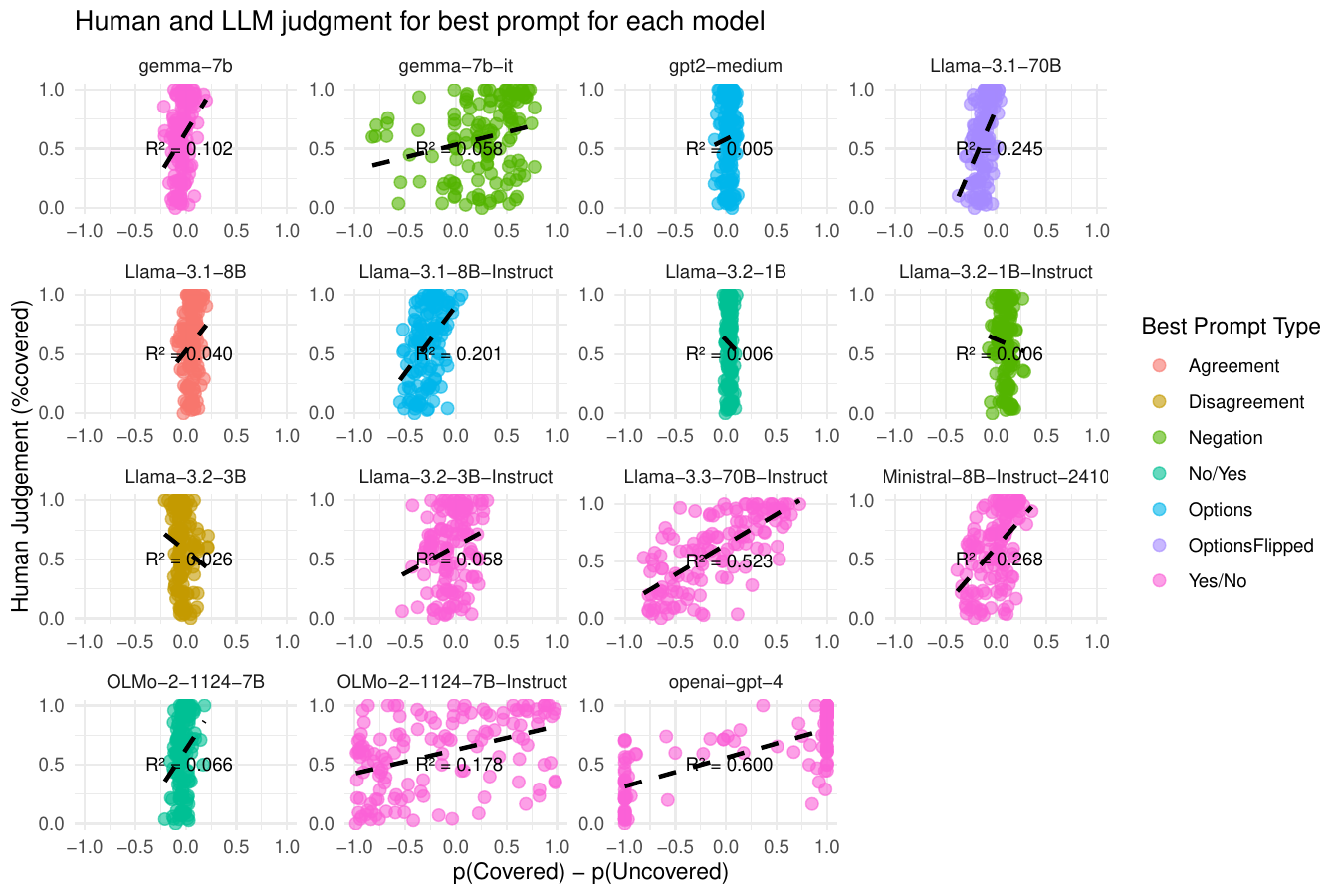}
    \caption{Each model's judgment probability versus human consensus with its prompt type reporting strongest correlation to human consensus. Dotted lines and the corresponding $R^2$ are best-fit lines.}
    \label{fig:best_prompt_for_each_model}
    \Description{Scatterplots of Human judgment variable and LLM judgment variables for the question variant with the best correlattion for each model. Each scatter plot visualizes values for one specific question variant, and also shows the best-fit line for Instruct LLM with the corresponding $R^2$ value. The best correlation is for 7 different variant (other than Agreement with Negation and Disagreement with Negation). 6 of 15 best correlated are for Yes/No variant.}
\end{figure*}

In \cref{sec:results-analysis}, we discuss LLMs' sensitivity to surface form and the lack of strong, consistent correlation to human judgment.
We provide additional detail on our analyses: correlation between human and \texttt{GPT-4} judgment across all prompt types in \cref{fig:gpt-human} and correlation between human and LLM judgment for each model's prompt type with the highest $R^2$ value in \cref{fig:best_prompt_for_each_model}.
The additional figures echo our findings.
\Cref{fig:gpt-human} illustrates sensitivity to trivial variation in input and response incongruence even in the model with the highest correlation to human judgment, \texttt{GPT-4}.
\Cref{fig:best_prompt_for_each_model} provides a detailed visualization of each model-prompt pair described in \cref{tab:strongest-correlation-by-model}.

\section{Linking hypothesis}
\label{sec:appendix-linking-hypothesis}
In \cref{sec:human-correlation}, we assume that our respective operationalizations of human and LLM judgment, namely the proportion of covered judgments \%\posjudg{Covered} in human responses and the probability difference between LLM covered and not covered judgments $\greendelta$, have a linear relationship. While the assumption is difficult to justify, we provide our attempt.

By definition, token probability $p(w\mid C) = f(C)$ in autoregressive language modeling \cite{radford2019language} represents the proportion of cases where the next token $w$ occurs given a model $f$ and number of environments with context $C$.
In our implementation, we consider this as a computational analogue of querying a human population and calculating the proportion of which that respond with one judgment, or, the proportion of human judgments in human responses.
This is the basis of our assumption that judgment probability as sum of token probabilities $p$(\posjudg{Covered}) = $p$(\postok{Yes}) + $p$(\postok{yes}) + $p$(\postok{YES}) and proportion of covered judgments have a linear relationship.

However, due to residual token probabilities that linger in LLM probability distributions, we are unable to represent LLM judgment with a single token probability, as a low $p$(\posjudg{Covered}) does not indicate a high $p$(\negjudg{NotCovered}).
We thus take the difference of the two probabilities, $\greendelta = p$(\posjudg{Covered}) $-$ $p$(\negjudg{NotCovered}) to represent LLM 
judgment that has a linear relationship with human majority judgment.
This allows for clear analysis, as the expectation value of a judgment $E(\greendelta ) = 0$, and $\Delta > 0$ yields the \posjudg{Covered} judgment and $\greendelta  < 0$ the \negjudg{NotCovered} judgment.

Based on our linking hypothesis, consider the ideal case where our two variables are perfectly correlated to each other with $R^2=1$. In such case where $\greendelta \in [-1, 1]$ and $p$(\posjudg{Covered}) $\in [0, 1]$, we predict that the proportion of human covered judgments is $0$ when probability difference is $-1$ since $p($\posjudg{Covered}$) = 0$, $p($\negjudg{NotCovered}$) = 1$. It follows that proportion of human covered judgments is 1 when probability difference is $1$; and 0.5 when probability difference $0$. The best fit line would then have $m = 0.5$, $b = 0.5$. Here, all of the variance across human judgment is explained by $\greendelta$.

Other metrics may be used to perform the same experiments, e.g. with normalization with the sum of relevant judgment tokens, disregarding residual token probabilities $p$(\textsf{\textcolor{gray}{other}}).

We outline $R^2$ measurements across 4 different formulas to represent LLM judgment in \cref{tab:operationalizations}. The four formulations differ, but as can be seen in the table, this does not affect the correlation values.\\
Let us define $\Delta = p(\text{\posjudg{Covered}}) - p(\text{\negjudg{NotCovered}})$ and $\Sigma = p(\text{\posjudg{Covered}}) + p(\text{\negjudg{NotCovered}})$.
Then,
$R^2_{\text{diff}}$ is with judgment as difference $\Delta$,
$R^2_{\text{covered}}$ is with judgment as just the covered judgment $p(\text{\posjudg{Covered}})$,
$R^2_{\text{rel}}$ is with judgment as relative quantity $p(\text{\posjudg{Covered}})/\Sigma$,
and $R^2_{\text{norm}}$ is with judgment as normalized difference $\Delta/\Sigma$.

We observe that the change in operationalization results in no significant change in $R^2$ values. In \cref{fig:best_prompt_for_each_model} you can see the lack of effect of the scale used for the judgment and the $R^2$. For contrast, between \texttt{Llama-8B-Inst} and \texttt{OLMo}. They have different scales, but similar correlations. 

\begin{table}[t!]
\centering
\small
\begin{tabular}{@{~}lHrrrr|lHrrrr@{~}}
\toprule
\textbf{Model} & \textbf{Var.} & $\bm{R^2_{\text{diff}}}$ & $\bm{R^2_{\text{cov}}}$ & $\bm{R^2_{\text{rel}}}$ & $\bm{R^2_{\text{norm}}}$ & \textbf{Model} & \textbf{Var.} & $\bm{R^2_{\text{diff}}}$ & $\bm{R^2_{\text{cov}}}$ & $\bm{R^2_{\text{rel}}}$ & $\bm{R^2_{\text{norm}}}$ \\
\midrule
Llama-70B   & Opt. F.     & 0.25 & 0.21 & 0.26 & 0.26 & Ministral & Y/N     & 0.27 & 0.25 & 0.27 & 0.27 \\
\quad +Inst & Y/N         & 0.52 & 0.50 & 0.53 & 0.53 & OLMo-7B   & N/Y        & 0.07 & 0.01 & 0.06 & 0.06 \\
Llama-8B    & Agr/Dis.    & 0.04 & 0.02 & 0.04 & 0.04 & \quad +Inst & Y/N        & 0.18 & 0.17 & 0.18 & 0.18 \\
\quad +Inst & Options     & 0.20 & 0.19 & 0.20 & 0.20 & gemma-7b  & Y/N  & 0.09 & 0.07 & 0.11 & 0.11 \\
Llama-3B    & Disagr.     & 0.03 & 0.02 & 0.03 & 0.03 & \quad +it   & Neg.       & 0.06 & 0.06 & 0.06 & 0.06 \\
\quad +Inst & Y/N         & 0.06 & 0.06 & 0.06 & 0.06 & GPT-4     & Y/N        & 0.60 & 0.60 & 0.60 & 0.60 \\
Llama-1B    & N/Y         & 0.01 & 0.00 & 0.01 & 0.01 & GPT-2-medium   & Options     & 0.01 & 0.01 & 0.01 & 0.00 \\
\quad +Inst & Neg.     & 0.01 & 0.00 & 0.00 & 0.00 &           &            &      &      &      &      \\
\bottomrule
\end{tabular}
\caption{$R^2$ values for all models and their highest human correlation. Each column represents an operationalization of LLM judgment.}

\label{tab:operationalizations}
\end{table}

\section{Limitations}
\label{sec:appendix-limitations}
Legal interpretation and `ordinary meaning' are complex topics of theory and practice in the legal field. Our work only looks at a specific aspect of LLM usage, posing direct queries for ascertaining ordinary meaning with a binary-choice QA task. This does not represent other mechanisms of using LMs for legal interpretation, such as producing arguments for and against an interpretation \cite{waldon_llms}, or eliciting examples \cite{almeman-etal-2024-wordnet}. Another concern is focused on different ways of eliciting such responses; cues are necessary to get consistent answers, but such cues can also constrain and distort the output to some extent as discussed in \citet{rottger-etal-2024-political}.

We use data from a previous study of consensus in legal interpretation. The authors of that study make no claims as to the overall representativeness of their experimental stimuli to questions that come up in day-to-day contractual interpretation. They also explicitly discuss the role of researcher subjectivity in the constructing the stimuli. We do not report any representativeness or coverage information for the data. Hence, despite the diversity of scenarios compared to other related works, it is currently unclear how representative it is of LLM use for legal interpretation in practice.

Our evaluation utilizes a small sample of 138 unique items. This is a small dataset for testing the generalization of language model judgments for the task.

We attempt to use question variants in a controlled manner to investigate how such variation affects judgment. However, the interplay between question variants and LLM judgment may not be easily disambiguated. LLMs' ability to understand and follow task instructions cannot be guaranteed, especially without targeted post-training~\cite{tamkin2023task}.

Our models are not chosen based on their attested performance on natural language benchmarks. Additionally, none of the models evaluated are considered ``reasoning'' or ``thinking'' models.

While we attempt to include models with varying size, architecture, and training recipe, we do not present our findings to be comprehensive or conclusive. Larger and more recent models with `reasoning', `retrieval' and/or search methods may be able to provide stable, reliable sources of human-like legal interpretation.
We also do not use ``chain-of-thought'' or other prompting or in-context learning methods meant to elicit or induce intermediate or reasoning steps. These have been shown to improve model performance in many tasks. Hence, our results may represent a lower estimation of model ability.

We use first token probability as the model response; this has advantages both in computing resources and analysis, but provides a limited representation of model output \cite{hu-levy-2023-prompting} but is effective for larger models \cite{song2025language}. Additionally, the first token probability has some known drawbacks \cite{wang2024look}, especially for instruction-tuned models, which we take additional steps to mitigate. Alternatively, non-instruction-tuned models are less likely to provide appropriate responses to our QA formulation.

Due to the complexity of responding to negation and interpreting it from the first token, we check the text to ascertain the correct polarity (\posjudg{Covered} or \posjudg{NotCovered}) to use for the judgments. However, this does not guarantee that all scenarios get captured under the polarity judgment. 
For distributional judgment, we collate the probabilities to only three categories, in which the tail of the probability distribution is reduced and represented as `Other'. This approximation reduces the precision and validity of the distance metric. 
Our correlation analysis uses a specific transformation of model responses and compares it to the \posjudg{Covered} proportion. We did not perform targeted validation for establishing our linking hypothesis in connecting human responses and LLM responses for legal interpretation. 

We did not perform checks for data contamination \cite{sainz-etal-2023-nlp} in the language models we have used. Data contamination of the published materials and reference policies used in the previous study \cite{waldon_vague_contracts_2023} could have influenced the models we studied. We catalog the cutoff and release dates in \Cref{sec:appendix-contamination}.

\end{document}